%% file: main.tex
\pdfoutput=1

\documentclass[11pt]{article}

\usepackage[final]{acl}

\usepackage{times}
\usepackage{latexsym}

\usepackage[T1]{fontenc}

\usepackage[utf8]{inputenc}

\usepackage{microtype}

\usepackage{multirow}
\usepackage{graphicx}
\usepackage{subfig}
\usepackage{algorithm}
\usepackage{algpseudocode}
\usepackage{amsmath}

\usepackage{inconsolata}

\newcommand{\namefull}{Batched Attention-optimized Speculative Sampling}
\newcommand{\nameshort}{BASS}

\newcommand{\ours}{BASS (ours)}
\newcommand{\reg}{RD (DS)}
\newcommand{\vllm}{RD (vLLM)}

\usepackage{multirow}
\usepackage{colortbl}

\newcommand{\skgouda}[1]{}
\newcommand{\myshang}[1]{}

\title{\nameshort: \namefull}

\author{Haifeng Qian$^{1,*}$ \quad Sujan Kumar Gonugondla$^{1,*}$ \quad Sungsoo Ha$^{2}$ \quad Mingyue Shang$^{1}$ \\ \textbf{Sanjay Krishna Gouda}$^{1}$ \quad \textbf{Ramesh Nallapati}$^{3}$ \quad \textbf{Sudipta Sengupta}$^{1}$ \quad \textbf{Xiaofei Ma}$^{1}$ \\ \textbf{Anoop Deoras}$^{1}$}

\begin{document}
\maketitle

\def\thefootnote{*}\footnotetext{Equal contribution.
$^{1}$AWS AI Labs.
$^{2}$AWS NGDE.
$^{3}$Amazon AGI (work done at AWS).
Correspondence to: Haifeng Qian <qianhf@amazon.com>, Sujan Kumar Gonugondla <gsujan@amazon.com>.}
\def\thefootnote{\arabic{footnote}}

\begin{abstract}
Speculative decoding has emerged as a powerful method to improve latency and throughput in hosting large language models.
However, most existing implementations focus on generating a single sequence.
Real-world generative AI applications often require multiple responses and how to perform speculative decoding in a batched setting while preserving its latency benefits poses non-trivial challenges.
This paper describes a system of batched speculative decoding that sets a new state of the art in multi-sequence generation latency and that demonstrates superior GPU utilization as well as quality of generations within a time budget.
For example, for a 7.8B-size model on a single A100 GPU and with a batch size of 8, each sequence is generated at an average speed of 5.8ms per token, the overall throughput being 1.1K tokens per second.
These results represent state-of-the-art latency and a 2.15$\times$ speed-up over optimized regular decoding.
Within a time budget that regular decoding does not finish, our system is able to generate sequences with HumanEval Pass@First of 43\% and Pass@All of 61\%, far exceeding what's feasible with single-sequence speculative decoding.
Our peak GPU utilization during decoding reaches as high as 15.8\%, more than 3$\times$ the highest of that of regular decoding and around 10$\times$ of single-sequence speculative decoding.
\end{abstract}

\input{intro}
\input{background}
\input{method}
\input{results}
\input{related}

\section{Conclusion}

This paper presents \namefull{} (\nameshort{}), a system that advances the state of the art in fast multi-sequence generation by LLMs.
By addressing the unique challenges of extending speculative decoding to batched inference without sacrificing latency, we demonstrate superior latency, GPU utilization as well as accuracy of generations within a time limit.

\input{ethic}

\bibliography{custom}

\input{appendix}

\end{document}

%% file: intro.tex
\section{Introduction}

In recent years, generative large language models (LLMs) have rapidly gained popularity due to their ability to generalize across a wide variety of tasks. These models are increasingly deployed commercially for applications such as coding assistants, writing aids, conversational agents, search and summarization tools and more. The accuracy performance of LLMs has been shown to scale with model size, with larger models demonstrating improved capabilities \cite{scaling_laws}. However, this improvement comes at the cost of greater latency during inference and increased computational requirements.

\begin{figure}[t]
    \centering
        \includegraphics[width=\linewidth]{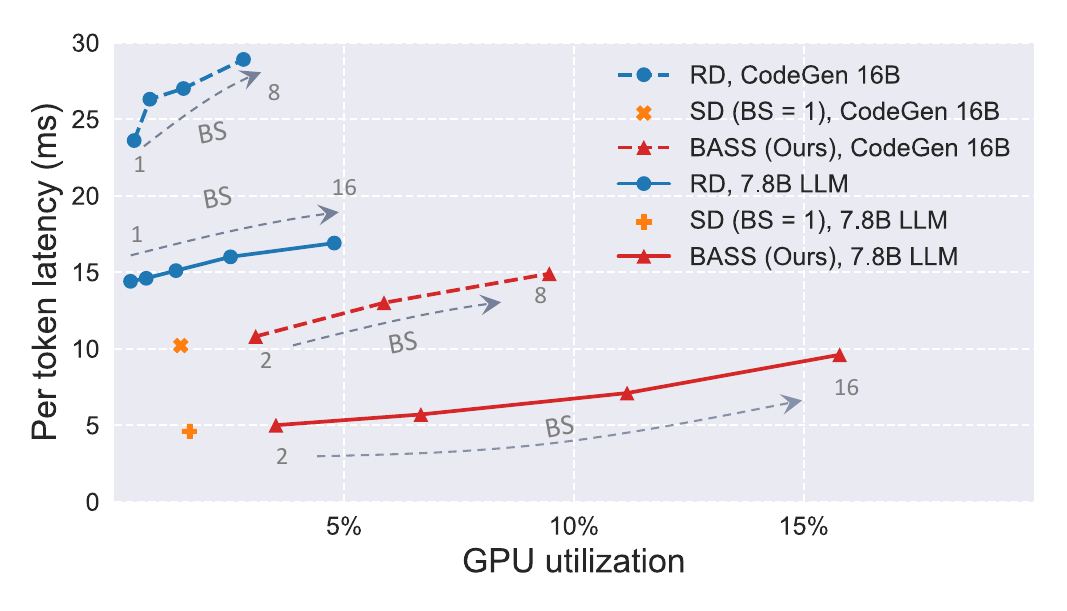}
        \caption{Comparing latency and GPU utilization of auto-regressive regular decoding (RD), single-sequence speculative decoding (SD) and our \nameshort{} method on two models. RD and \nameshort{} are measured with exponentially increasing batch sizes (BS).}
    \label{fig:util}
\end{figure}

Most popular LLMs are transformer-based decoder models. The inference speed of these models is often limited by memory bandwidth on typical hardware like GPUs.  This is because GPUs tend to have much higher compute throughput relative to memory bandwidth. The auto-regressive decoding process of these models, where each output token is generated sequentially conditioned on previous tokens, means the entire model parameters need to be fetched from memory for each generated token. This sequential nature prevents parallelization during inference, resulting in under-utilization of available compute resources. For example, for both models in Figure~\ref{fig:util}, single-sequence regular decoding utilizes only 0.4\% of GPU FLOPS.

To improve GPU utilization, batching multiple sequences is often employed to amortize the memory I/O costs across a batch and thereby utilize more FLOPS per memory I/O. However, large batch sizes are needed to effective utilize GPU compute, resulting in higher latency for individual sequences that are batched as well as bigger memory footprints.  With larger model sizes, memory bottleneck becomes a challenge and limits allowable batch sizes.
In Figure~\ref{fig:util} for example, the highest GPU utilization by batched regular coding is only 4.8\% before going out-of-memory.

Speculative decoding has emerged as an effective approach to improve latency of LLMs by increasing GPU utilization. The key idea is to draft a few tokens (typically by using a smaller LLM) and verify their correctness with the main LLM. By processing the draft tokens in parallel, speculative decoding amortizes the memory I/O of model parameters across the tokens. Despite its advantages, speculative decoding has limitations: It processes a single sequence at a time, restricting the parallelism to the number of draft tokens. This caps the potential GPU utilization improvements.

To address this, we present \namefull{} (\nameshort{}) -- a parallel speculative decoder that handles multiple sequences simultaneously.
\nameshort{} increases GPU utilization by parallelism across both the batch dimension and the draft-token dimension.
We implement customized CUDA kernels to handle ragged tensors during attention calculation, which are a challenge posed by batched speculative decoding, and design a heuristic to dynamically adjust draft length for each step.
As illustrated in Figure~\ref{fig:util}, \nameshort{} achieves latency and GPU utilization that are substantially improved from prior regular and speculative decoders.
By comprehensive experiments on three different models including CodeGen and OPT, we study these trends as well as accuracy benefits of \nameshort{}.
We study the impact of draft model design on overall system performance as well as that of algorithmic choices in attention kernels and draft lengths.
\nameshort{} is applicable to both batch generation from a same prompt and batch generation from a set of different prompts.

%% file: background.tex
\section{Background}

\subsection{Inference with LLMs}

\begin{figure}
    \centering
        \includegraphics[width=\linewidth]{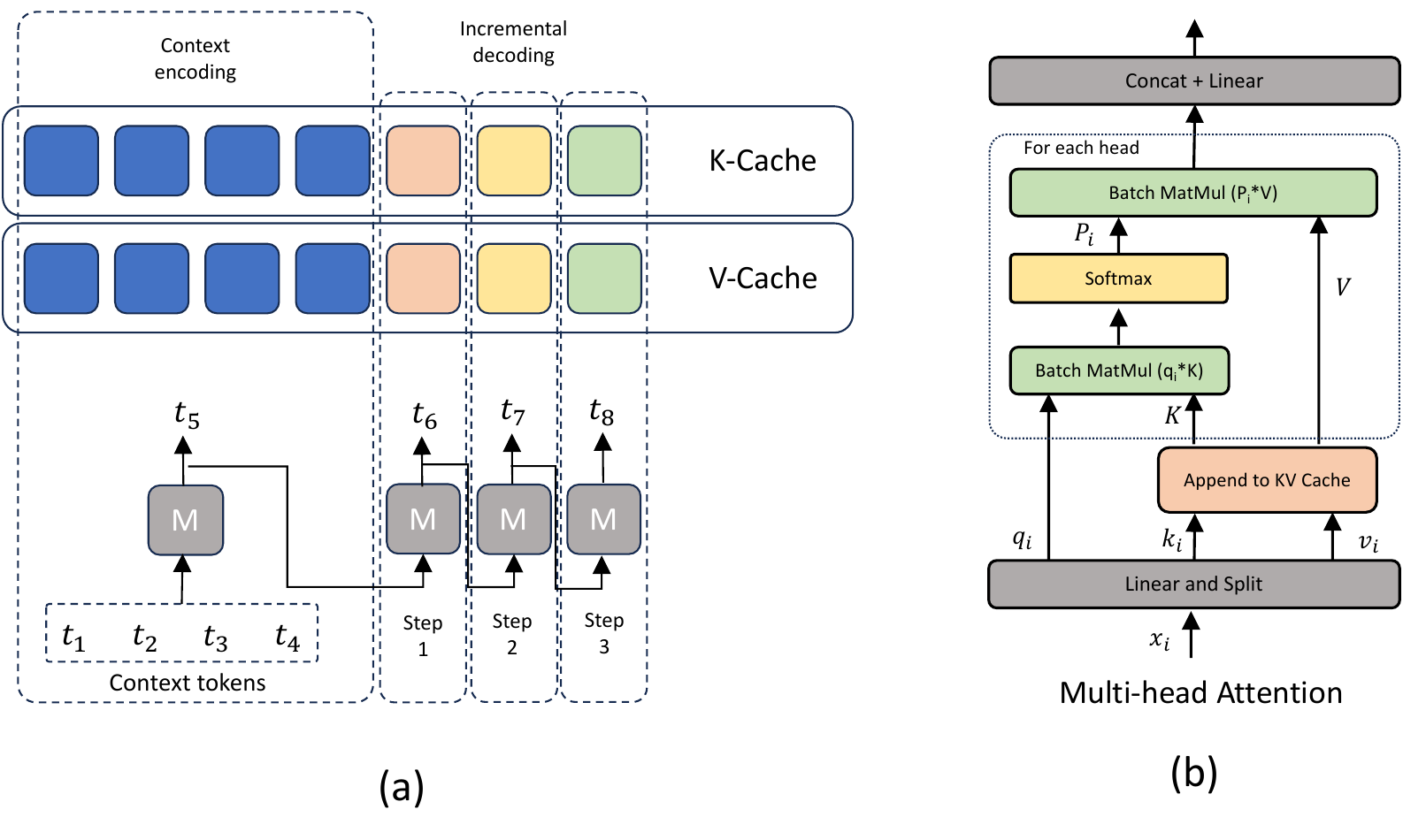}
        \caption{(a) Inference steps in regular decoding of an LLM. (b) Operations in multi-head attention.}
    \label{fig:inference-attention}
\end{figure}

This paper focuses on transformer-based \cite{attention_all_you_need} decoder-only generative LLMs. The standard inference of these LLMs can be divided into two phases:
(a) \textit{context encoding (prefill) phase} where the input prompt is processed in parallel to encode contextual information, and
(b) \textit{incremental decoding phase} where the model auto-regressively generates output tokens one by one based on the encoded context (Figure~\ref{fig:inference-attention}(a)).

Consider a decoder-only transformer \cite{radford2019language} architecture with alternatively stacked feed-forward network layers and attention layers. In the attention mechanism (Figure \ref{fig:inference-attention}(b)), key ($k_i$), value ($v_i$), and query ($q_i$) vectors are first computed by projecting the token embeddings for each position $i$. The queries are then used to calculate the relevance of the current token with past positions, by estimating their correlation with the past keys.

During the context encoding phase, all prompt tokens are processed in parallel. The attention keys and values for all context tokens ($K$ and $V$ tensors) are cached during this phase. This phase exhibits high GPU utilization.

The incremental decoding phase is bottlenecked by memory I/O from repeated fetching of model parameters and the KV cache as each output token is decoded.
This phase is typically the dominant portion of inference time and exhibits low GPU utilization.
It is the focus of our optimizations.

\subsection{Speculative decoding}
\label{sec:background-spec}
\begin{figure}
    \centering
        \includegraphics[width=\linewidth]{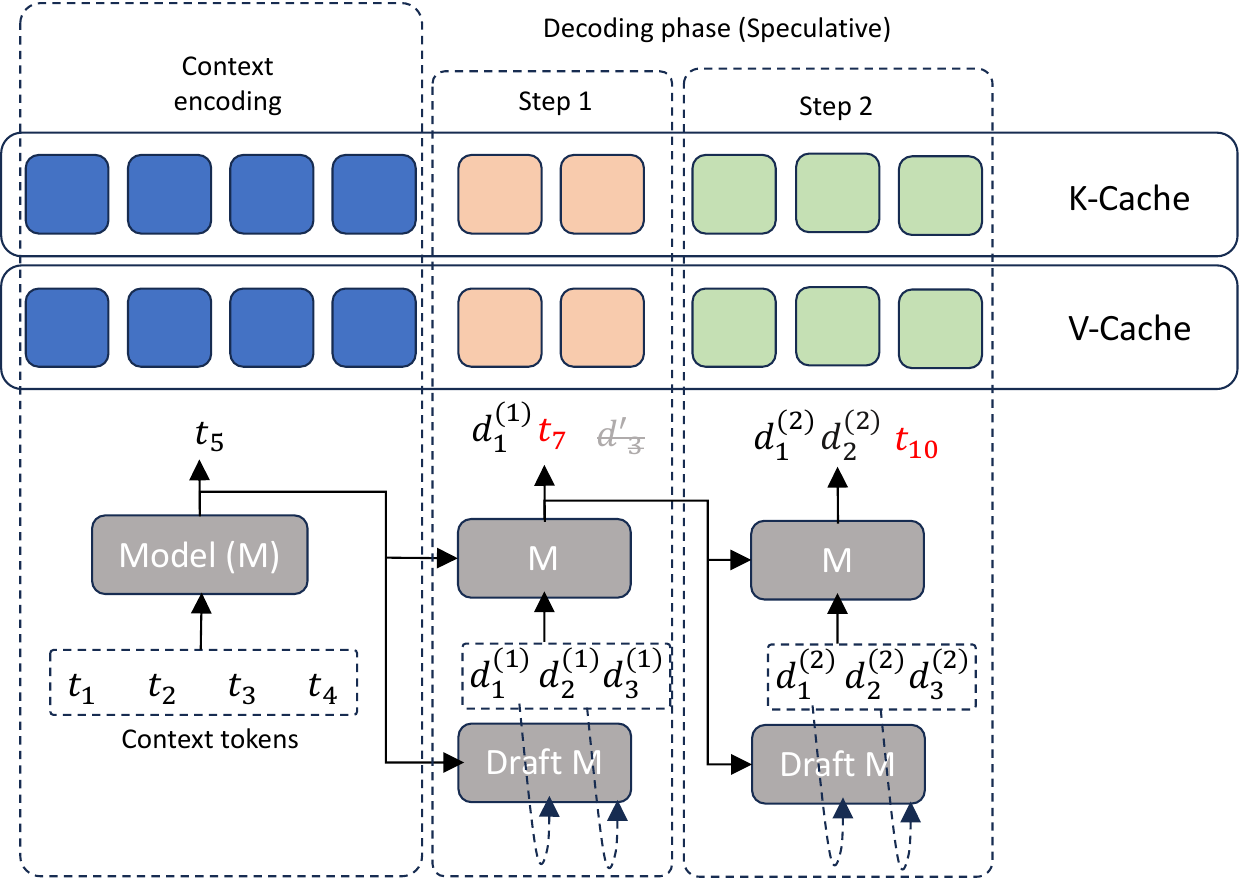}
        \caption{Standard speculative decoding. The draft model (Draft M) generates $k$ draft tokens auto-regressively, which are then processed by the main model (M) in parallel to verify correctness.}
    \label{fig:speculative}
\end{figure}

Speculative decoding \cite{stern2018blockwise, xia2022speculative, leviathan2023fast, chen2023accelerating} is a popular technique to reduce latency of LLM inference. As illustrated in Figure \ref{fig:speculative}, the idea is to use a small draft model to generate $k$ draft tokens. They are then processed by the main model as incremental context encoding to decide which draft tokens to accept.

The number of draft tokens to accept is decided based on the probabilities of generating these tokens according to the main and draft models. In the case of rejection, a corrected token is sampled from the outputs of the main model. Overall, this decision process is stochastic and is designed to be equivalent to sampling from the main model’s distribution \cite{leviathan2023fast, chen2023accelerating}. For example, we may accept the first five tokens, correct the sixth token, and then generate new draft tokens from the seventh position.

Latency benefit comes from the higher GPU utilization associated with incremental context encoding rather than auto-regressive decoding. More than 2$\times$ latency reduction has been reported in  literature \cite{leviathan2023fast,chen2023accelerating}.

\subsubsection{Limitations of speculative decoding}
\label{sec:spec-limitations}
A major limitation of speculative decoding is that batch size of the main model's generation
is preferably just 1, which is the setting in most existing works. It is straightforward to see why. In a naive implementation with batch size more than 1, we stop accepting tokens at the first reject position in the batch and hence lose some latency benefit. For illustration, let’s make a simplistic assumption that each token generated by the draft model has an independent chance $p$ of getting accepted, then the number of output tokens per draft has a geometric distribution with an expected value of $1/(1-p)$. For example, if $p=80$\%, then on average the decoding process moves forward by 5 tokens per draft. With a batch size of $b$, the probability of acceptance per position becomes $p^b$. For a batch size of five as an example, the probability of acceptance per position becomes 33\% and on average the decoding process moves forward by merely 1.5 tokens per draft, and we have lost most, if not all, of the latency benefit.

To retain the latency benefit with $b>1$, we need to accept variable numbers of draft tokens across a batch.
This poses challenges that no prior work has addressed efficiently and the existing systems remain single-sequence inference.

%% file: method.tex
\section{\namefull{}}
\label{sec:method}

\namefull{} (\nameshort{}) extends speculative decoding by enabling batch processing across multiple sequences. While speculative sampling improves GPU utilization for a single sequence, parallelism is limited to the small number of draft tokens per sequence (typically 5-10). Batching sequences with speculative decoding can further maximize GPU usage. To fully realize these benefits, specialized tensor manipulations and CUDA kernels are required.

\subsection{Challenges with batching}

One challenge with batched speculative decoding stems from the uncertainty in the numbers of acceptable draft tokens for each sequence, which vary across the batch.
LLM inference kernels are designed to handle regular shaped tensors, primarily driven by CUDA restrictions.
If we enforce a uniform sequence length across the batch, that would result in less accepted draft tokens and in diminishing benefits as discussed in Section~\ref{sec:spec-limitations}. 

In order to maintain the performance gains of speculative sampling, we need to be able to accept variable numbers of tokens across the batch.
This will result in variable sequence lengths across the batch and consequently ragged-shape $K$ and $V$ tensors.
In particular, this will affect the computations in the attention layer where we may not be able to batch the operations into a single kernel.
Handling this effectively needs careful design considerations.

Another challenge is the choice of draft lengths.
Although prior experimental systems tend to use a fixed draft length, real-life deployment requires adaptive draft lengths because different prompts may lead to different degrees of alignment between draft and main models during generation and hence different optimal draft lengths.
For efficient incremental context encoding by the main model, we need to choose a uniform draft length across the batch at each step, and this decision needs to balance the needs of the multiple sequences.

\subsection{Proposed method}

\begin{figure}
    \centering
        \includegraphics[width=\linewidth]{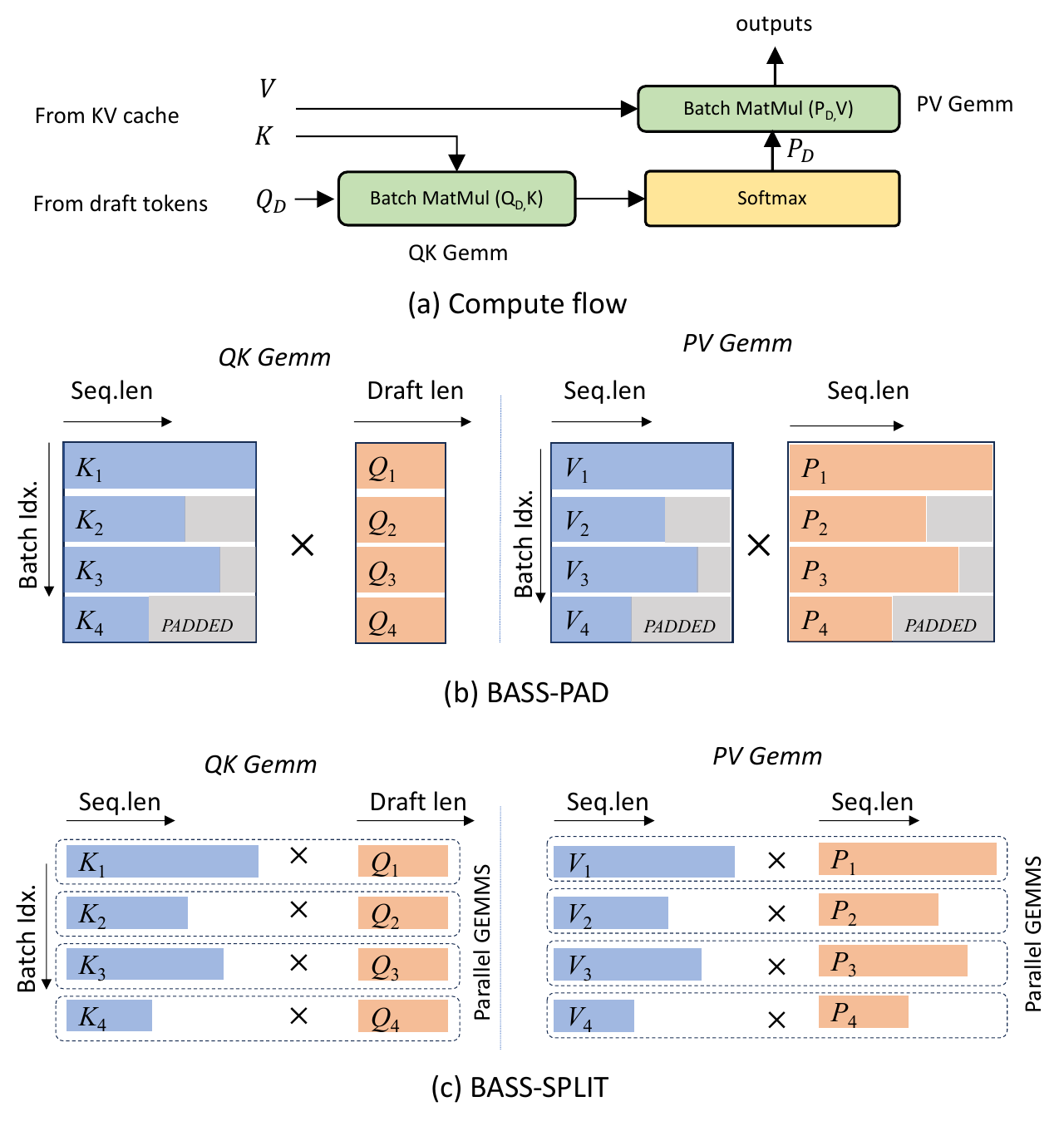}
        \caption{Attention calculation in \nameshort{}: (a) Attention compute flow, (b) \nameshort{}-PAD launches one kernel for QK GEMM and one kernel for PV GEMM by padding the $K$, $V$ and $P$ tensors to the maximum sequence length across the batch, and (c) \nameshort{}-SPLIT launches one kernel per sequence and thereby accommodates variable sequence lengths.}
    \label{fig:method}
\end{figure}

The attention computation flow, illustrated in Figure~\ref{fig:method}(a), involves a GEMM\footnote{General matrix multiplication (GEMM).} operation between the query tensor $Q$ of the current tokens and the key tensor $K$ of all tokens processed so far, a softmax operation that outputs $P =\mathrm{softmax} (Q^TK/c)$, and another GEMM operation between $P$ and value tensor $V$ of all tokens processed so far.
There are three ragged tensors, $K$, $V$ and $P$.
We propose two variants: \nameshort{}-PAD and \nameshort{}-SPLIT as illustrated in Figure~\ref{fig:method}(b) and Figure~\ref{fig:method}(c) respectively.
They implement the two GEMMs differently and share the same softmax operation: we simply launch separate softmax kernels, one for each sequence.

In the \nameshort{}-PAD approach, we pad the $K$, $V$ and $P$ tensors along the sequence length dimension to match the longest sequence of the batch and use the padded tensors for computations.
We assign zero probabilities for the padded tokens in $P$.
\nameshort{}-PAD does not incur additional cost of launching kernels, but we waste some compute to perform dummy computations with the padded elements.

The \nameshort{}-SPLIT approach is derived from the insight that attention operation is not associated with any model parameters and therefore applying batching to attention has no benefit of reducing the amount of GPU memory reads.
In this approach, we break up the KQ GEMM and the PV GEMM into smaller per-sequence kernels, so that we can handle the variable sequence-length dimension in $K$, $V$ and $P$ tensors, as shown in Figure~\ref{fig:method}(c).
Note that these separate kernels can be launched and executed in parallel.
Since there is no sharing of memory reads among these separate kernels, the only extra cost that we pay is the cost of launching them.
\nameshort{}-SPLIT does not waste any compute.

\nameshort{}-PAD and \nameshort{}-SPLIT apply to both the main and the draft models and apply to both token generation and incremental context encoding. With either approach, we can now let each sequence proceed at its own pace according to its own reject points and let sequences in the batch have different lengths.

Note that other steps, including the feed-forward network, the KQV projection layer, and the projection layer at end of attention, all remain the same and remain regularly batched to share memory reads of model parameters.

The comparison between \nameshort{}-PAD and \nameshort{}-SPLIT depends on the application scenario.
We find \nameshort{}-PAD more attractive when the sequence lengths are closer to each other and \nameshort{}-SPLIT more favorable when the sequence lengths across batch vary by a large margin.
Most results in Section~\ref{sec:results} are based on \nameshort{}-PAD and their comparison results are in Section~\ref{sec:ablation}.

\begin{algorithm}
\caption{A heuristic to adjust draft length}\label{alg:draftlength}
\begin{algorithmic}
\State $l_\text{draft} \gets l_0$ 
\State $s \gets $ 0
\For{each speculative decoding step}
\State $x_1,\cdots,x_b \gets$ numbers of accepted tokens 
\If{$\max\left( x_1,\cdots,x_b \right) = l_\text{draft}$}
    \State $l_\text{draft} \gets \min\left(l_\text{draft} + l_\text{incre},l_\text{limit}\right)$
    \State $s \gets $ 0
\Else
    \State $l_\text{draft} \gets l_\text{draft} - \lceil l_\text{draft}/l_\text{mod} \rceil - s$
    \State $l_\text{draft} \gets \max\left(1,x_1,\cdots,x_b,l_\text{draft}\right)$
    \State $s \gets $ 1
\EndIf
\EndFor
\end{algorithmic}
\end{algorithm}

Algorithm~\ref{alg:draftlength} describes the heuristic that we use to dynamically adjust draft lengths.
The rationale is to increase draft length when at least one sequence has accepted all draft tokens in the last speculative decoding step and to decrease it otherwise.
The speed of decrease is larger when the current draft length is larger and when it is a consecutive step of decrease.
However, the next draft length is never decreased to be less than the max number of accepted tokens in the batch.
We empirically chose the parameters of $l_0=7$, $l_\text{incre}=2$, $l_\text{mod}=10$ and $l_\text{limit}=32$.
Comparisons against constant draft lengths are in Section~\ref{sec:ablation}.

The degree of alignment between draft and main models varies across prompts, across different sequences from the same prompt, and also within the same sequence. When generating commonly used sentences or boilerplate code for example, the alignment tends to be strong and the optimal draft length is long. When generating uncommon sentences or novel code segments, the alignment tends to be weak and the optimal draft length is short. The effect of Algorithm 1 versus a fixed draft length is to dynamically get closer to the optimal draft length in both scenarios, so that the system generates longer drafts where possible yet does not waste compute to generate throw-away tokens.

%% file: results.tex
\section{Experiments}
\label{sec:results}
In this section we demonstrate the benefits of \nameshort{} over batched auto-regressive regular decoding (RD) and single-sequence speculative decoding.

\myshang{If space allows, it's better to add a paragraph here talking about the structure of this section. e.g. in this section, we first describe the experimental settings and then we report the results on different down-stream tasks including xxx, xxx and xxxx. Our experiments shows (high-level observations.)}
\begin{table*}[th]
\small
\centering
\begin{tabular}{cclcrrrrrr}
\hline
\multirow{2}{*}{Prec.} & \multirow{2}{*}{Batch} & \multirow{2}{*}{Method} & \multirow{2}{*}{ROUGE-2} & \multicolumn{6}{c}{Mean per-token latency \& Speedup}\\
\cline{5-10}
&&&& \multicolumn{2}{c}{First} & \multicolumn{2}{c}{Last} & \multicolumn{2}{c}{All} \\
\hline
\multirow{9}{*}{FP16}
& \multirow{3}{*}{1} & \reg{}& 0.086 & 23.4 ms & 1$\times$ & 23.4 ms & 1$\times$ & 23.4 ms & 1$\times$ \\
&  & \vllm{} & 0.083 & 24.0 ms & 0.98$\times$ & 24.0 ms & 0.98$\times$ & 24.0 ms & 0.98$\times$ \\
&   & \cellcolor{gray!20}\ours{}    
    & \cellcolor{gray!20}0.084 
    & \cellcolor{gray!20}10.8 ms 
    & \cellcolor{gray!20}2.16$\times$ 
    & \cellcolor{gray!20}10.8 ms 
    & \cellcolor{gray!20}2.16$\times$ 
    & \cellcolor{gray!20}10.8 ms 
    & \cellcolor{gray!20}2.16$\times$ \\
\cline{2-10}
& \multirow{3}{*}{2} & \reg{}& 0.085 & 25.9 ms & 1$\times$ & 25.9 ms & 1$\times$ & 25.9 ms & 1$\times$ \\
&  & \vllm{} & 0.084 & 23.9 ms & 1.08$\times$ & 23.9 ms & 1.08$\times$ & 23.9 ms & 1.08$\times$ \\
&  & \cellcolor{gray!20}\ours{}    
    & \cellcolor{gray!20}0.084 
    & \cellcolor{gray!20}9.4 ms 
    & \cellcolor{gray!20}2.74$\times$ 
    & \cellcolor{gray!20}12.6 ms 
    & \cellcolor{gray!20}2.05$\times$ 
    & \cellcolor{gray!20}11.0 ms 
    & \cellcolor{gray!20}2.34$\times$ \\
\cline{2-10}
& \multirow{3}{*}{4} & \reg{}& 0.085 & 27.0 ms & 1$\times$ & 27.0 ms & 1$\times$ & 27.0 ms & 1$\times$ \\
&  & \vllm{} & 0.084 & 24.3 ms & 1.11$\times$ & 24.3 ms & 1.11$\times$ & 24.3 ms & 1.11$\times$ \\
&   & \cellcolor{gray!20}\ours{}    
    & \cellcolor{gray!20}0.084 
    & \cellcolor{gray!20}9.6 ms 
    & \cellcolor{gray!20}2.81$\times$ 
    & \cellcolor{gray!20}16.6 ms 
    & \cellcolor{gray!20}1.62$\times$ 
    & \cellcolor{gray!20}12.7 ms 
    & \cellcolor{gray!20}2.13$\times$ \\
\hline
\multirow{8}{*}{INT8}
& \multirow{2}{*}{1} & \reg{}& 0.085 & 17.4 ms & 1$\times$ & 17.4 ms & 1$\times$ & 17.4 ms & 1$\times$ \\
&  & \cellcolor{gray!20}\ours{}    & \cellcolor{gray!20}0.087 & \cellcolor{gray!20}8.5 ms & \cellcolor{gray!20}2.05$\times$ & \cellcolor{gray!20}8.5 ms & \cellcolor{gray!20}2.05$\times$ & \cellcolor{gray!20}8.5 ms & \cellcolor{gray!20}2.05$\times$ \\
\cline{2-10}
& \multirow{2}{*}{2} & \reg{}& 0.086 & 20.1 ms & 1$\times$ & 20.1 ms & 1$\times$ & 20.1 ms & 1$\times$ \\ 
&  
    & \cellcolor{gray!20}\ours{}    
    & \cellcolor{gray!20}0.087 
    & \cellcolor{gray!20}7.8 ms 
    & \cellcolor{gray!20}2.57$\times$ 
    & \cellcolor{gray!20}10.7 ms 
    & \cellcolor{gray!20}1.87$\times$ 
    & \cellcolor{gray!20}9.3 ms 
    & \cellcolor{gray!20}2.16$\times$ \\
\cline{2-10}
& \multirow{2}{*}{4} & \reg{}& 0.086 & 21.1 ms & 1$\times$ & 21.1 ms & 1$\times$ & 21.1 ms & 1$\times$ \\
&  
    & \cellcolor{gray!20}\ours{}    
    & \cellcolor{gray!20}0.087 
    & \cellcolor{gray!20}8.2 ms 
    & \cellcolor{gray!20}2.58$\times$ 
    & \cellcolor{gray!20}14.8 ms 
    & \cellcolor{gray!20}1.43$\times$ 
    & \cellcolor{gray!20}11.2 ms 
    & \cellcolor{gray!20}1.88$\times$ \\
\cline{2-10}
& \multirow{2}{*}{8} & \reg{}& 0.086 & 23.5 ms & 1$\times$ & 23.5 ms & 1$\times$ & 23.5 ms & 1$\times$ \\
&  
    & \cellcolor{gray!20}\ours{}    
    & \cellcolor{gray!20}0.087 
    & \cellcolor{gray!20}9.6 ms 
    & \cellcolor{gray!20}2.44$\times$ 
    & \cellcolor{gray!20}21.7 ms 
    & \cellcolor{gray!20}1.08$\times$ 
    & \cellcolor{gray!20}14.5 ms 
    & \cellcolor{gray!20}1.62$\times$ \\
\hline
\end{tabular}
\caption{OPT 13B accuracy and latency on XSum with auto-regressive regular decoding (RD) with DeepSpeed (DS) and vLLM, and \nameshort{}. Temperature is 0.2, nucleus top $p$ is 0.95, and draft model is OPT 125M.}
\label{tab:opt}
\end{table*}

\subsection{Setup}

\noindent {\bf Inference setup and CUDA kernels:}
All experiments throughout this paper are conducted on a single A100 GPU with 40GB memory.
All inference runs, except for the ``vLLM'' rows in Tables~\ref{tab:opt} and \ref{tab:codegen}, use a modified version of DeepSpeed\footnote{\url{https://github.com/microsoft/DeepSpeed}} (DS), including both regular decoding and \nameshort{} runs.
As can be seen in the tables, the regular decoding latencies are at the state of the art and comparable to those reported in \cite{aminabadi2022deepspeed,yao2022zeroquant} for both FP16 and INT8.
The vLLM runs use the latest vLLM version\footnote{\url{https://github.com/vllm-project/vllm}} (v0.3.0) and all sequences start immediately and hence result in the best possible latencies.
Our modifications to DeepSpeed include:
\begin{itemize}
\item Kernels for quantizing both weights and activations to INT8 for all linear layers. We use CUTLASS\footnote{\url{https://github.com/NVIDIA/cutlass}} INT8$\rightarrow$INT32 kernels for GEMM calls and modify them and other layers to fuse quantization and de-quantization operators.
\item Kernels for attention calculations to address the ragged-tensor challenge in batched speculative decoding without sacrificing latency, as discussed in Section~\ref{sec:method}.
\end{itemize}

\noindent {\bf Models and tasks:} We report experimental results on three main models with their respective draft models and tasks\footnote{While the three tasks are the scenario of batch generation from a same prompt, please note that \nameshort{} is also applicable to batch generation from a set of different prompts.}:
\begin{itemize}
\item
OPT 13B as the main model and OPT 125M or 350M as the draft model \cite{zhang2022opt}.
Following the same experimental setting as \cite{chen2023accelerating}, we use the XSum task with 11,334 test examples \cite{narayan2018don}, use 1-shot prompting, generate 128 tokens per sequence, and use ROUGE-2 \cite{lin2004rouge} as the metric to verify accuracy.
\item
CodeGen-Mono 16B as the main model and CodeGen-Mono 350M as the draft model.
We use the HumanEval task with 164 examples and use the Pass@$K$ accuracy metric \cite{chen2021evaluating},
and we generate 256 tokens for each sequence.
\item
A 7.8B model trained on text and code as the main model and one of three draft models with sizes 310M, 510M and 1B.
We again use the HumanEval task.
\end{itemize}

\noindent{\bf Latency metrics:} We use the metric of per-token latency for each generated sequence \footnote{It is important to note that we do not divide latency by batch size, which was done in some papers, e.g., \cite{su2023synergy}. Fundamentally, our definition is a latency metric while the definition in \cite{su2023synergy} is a throughput metric. With our definition, per token latency increases as the batch size increases because the amount of FLOPS during the per token latency is multiplied by batch size, and this applies to both regular decoding and speculative decoding.}.
For regular decoding, this value is the same across a batch.
For speculative decoding, it varies within a batch and therefore we report three metrics: per-token latency of the first finished sequence in a batch, per-token latency of the last finished sequence in a batch, and per-token latency averaged across a batch, where each of the three is then averaged across examples in a task dataset.

\subsection{Performance on summarization}

Table~\ref{tab:opt} shows the accuracy and latency results of the OPT 13B model on the summarization task of the XSum dataset, with OPT 125M as the draft model.
As expected, the results suggest neutral accuracy between regular decoding and speculative decoding, while speculative decoding provides up to 2.81$\times$ speed up for finishing the first sequence and up to 2.34$\times$ speed up on average for all sequences.
In a real-life application, and particularly in the scenario of generating multiple sequences for the same prompt, we can respond to the user as soon as the first sequence finishes while the other additional recommendations continue to generate.
Therefore, speeding up the first sequence by 2.05$\times$--2.81$\times$ implies a significant improvement in user-perceived latency.

The latency divergence between the first and last finished sequences increases with batch size. However, the last sequence latency is not as important in a real-life application because, when generating multiple sequences for the same prompt, we can simply choose a cut-off latency limit and return, e.g., five finished sequences out of a batch of eight.

\begin{table*}[t]
    \small
    \centering
    \begin{tabular}{cclcrrrrrr}
    \hline
    \multirow{2}{*}{Prec.} & \multirow{2}{*}{Batch} & \multirow{2}{*}{Method} & \multirow{2}{*}{Pass@Batch} & \multicolumn{6}{c}{Mean per-token latency \& Speedup}\\
    \cline{5-10}
    &&&& \multicolumn{2}{c}{First} & \multicolumn{2}{c}{Last} & \multicolumn{2}{c}{All} \\
    \hline
    \multirow{12}{*}{FP16}
    & \multirow{3}{*}{1} & \reg{}& 30.5\% & 23.6 ms & 1$\times$ & 23.6 ms & 1$\times$ & 23.6 ms & 1$\times$ \\
    &  & \vllm{} & 31.0\% & 26.7 ms & 0.88$\times$ & 26.7 ms & 0.88$\times$ & 26.7 ms & 0.88$\times$ \\
    &  
    & \cellcolor{gray!20}\ours{}    
    & \cellcolor{gray!20}30.5\% 
    & \cellcolor{gray!20}10.2 ms 
    & \cellcolor{gray!20}2.31$\times$ 
    & \cellcolor{gray!20}10.2 ms 
    & \cellcolor{gray!20}2.31$\times$ 
    & \cellcolor{gray!20}10.2 ms 
    & \cellcolor{gray!20}2.31$\times$ \\
    \cline{2-10}
    & \multirow{3}{*}{2} & \reg{}& 36.6\% & 26.3 ms & 1$\times$ & 26.3 ms & 1$\times$ & 26.3 ms & 1$\times$ \\
    &  & \vllm{} & 35.9\% & 28.2 ms & 0.93$\times$ & 28.2 ms & 0.93$\times$ & 28.2 ms & 0.93$\times$ \\
    &  
    & \cellcolor{gray!20}\ours{}    
    & \cellcolor{gray!20}36.0\% 
    & \cellcolor{gray!20}9.9 ms 
    & \cellcolor{gray!20}2.65$\times$ 
    & \cellcolor{gray!20}11.7 ms 
    & \cellcolor{gray!20}2.25$\times$ 
    & \cellcolor{gray!20}10.8 ms 
    & \cellcolor{gray!20}2.43$\times$ \\
    \cline{2-10}
    & \multirow{3}{*}{4} & \reg{}& 39.0\% & 27.0 ms & 1$\times$ & 27.0 ms & 1$\times$ & 27.0 ms & 1$\times$ \\
    &  & \vllm{} & 40.4\%  & 28.9 ms & 0.93$\times$ & 28.9 ms & 0.93$\times$ & 28.9 ms & 0.93$\times$ \\
    &  
    & \cellcolor{gray!20}\ours{}    
    & \cellcolor{gray!20}40.2\% 
    & \cellcolor{gray!20}10.8 ms 
    & \cellcolor{gray!20}2.50$\times$ 
    & \cellcolor{gray!20}15.6 ms 
    & \cellcolor{gray!20}1.73$\times$ 
    & \cellcolor{gray!20}13.0 ms 
    & \cellcolor{gray!20}2.07$\times$ \\
    \cline{2-10}
    & \multirow{3}{*}{8} & \reg{}& 42.7\% & 28.9 ms & 1$\times$ & 28.9 ms & 1$\times$ & 28.9 ms & 1$\times$ \\
    &  & \vllm{} &  45.1\% & 29.7 ms & 0.97$\times$ & 29.7 ms & 0.97$\times$ & 29.7 ms & 0.97$\times$ \\
    &  
    & \cellcolor{gray!20}\ours{}    
    & \cellcolor{gray!20}45.1\% 
    & \cellcolor{gray!20}11.5 ms 
    & \cellcolor{gray!20}2.51$\times$ 
    & \cellcolor{gray!20}19.4 ms 
    & \cellcolor{gray!20}1.49$\times$ 
    & \cellcolor{gray!20}14.9 ms 
    & \cellcolor{gray!20}1.94$\times$ \\
    \hline
    \multirow{8}{*}{INT8}
    & \multirow{2}{*}{1} & \reg{}& 32.3\% & 16.8 ms & 1$\times$ & 16.8 ms & 1$\times$ & 16.8 ms & 1$\times$ \\
    &  
    & \cellcolor{gray!20}\ours{}    
    & \cellcolor{gray!20}31.7\% 
    & \cellcolor{gray!20}9.3 ms 
    & \cellcolor{gray!20}1.82$\times$ 
    & \cellcolor{gray!20}9.3 ms 
    & \cellcolor{gray!20}1.82$\times$ 
    & \cellcolor{gray!20}9.3 ms 
    & \cellcolor{gray!20}1.82$\times$ \\
    \cline{2-10}
    & \multirow{2}{*}{2} & \reg{}& 36.6\% & 19.6 ms & 1$\times$ & 19.6 ms & 1$\times$ & 19.6 ms & 1$\times$ \\
    &  
    & \cellcolor{gray!20}\ours{}    
    & \cellcolor{gray!20}36.0\% 
    & \cellcolor{gray!20}9.3 ms 
    & \cellcolor{gray!20}2.11$\times$ 
    & \cellcolor{gray!20}10.9 ms 
    & \cellcolor{gray!20}1.79$\times$ 
    & \cellcolor{gray!20}10.1 ms 
    & \cellcolor{gray!20}1.94$\times$ \\
    \cline{2-10}
    & \multirow{2}{*}{4} & \reg{}& 38.4\% & 20.4 ms & 1$\times$ & 20.4 ms & 1$\times$ & 20.4 ms & 1$\times$ \\
    &  
    & \cellcolor{gray!20}\ours{}    
    & \cellcolor{gray!20}39.0\% 
    & \cellcolor{gray!20}9.8 ms 
    & \cellcolor{gray!20}2.07$\times$ 
    & \cellcolor{gray!20}13.2 ms 
    & \cellcolor{gray!20}1.54$\times$ 
    & \cellcolor{gray!20}11.2 ms 
    & \cellcolor{gray!20}1.81$\times$ \\
    \cline{2-10}
    & \multirow{2}{*}{8} & \reg{}& 44.5\% & 21.9 ms & 1$\times$ & 21.9 ms & 1$\times$ & 21.9 ms & 1$\times$ \\
     &  
     & \cellcolor{gray!20}\ours{}    
     & \cellcolor{gray!20}42.7\% 
     & \cellcolor{gray!20}11.1 ms 
     & \cellcolor{gray!20}1.98$\times$ 
     & \cellcolor{gray!20}18.8 ms 
     & \cellcolor{gray!20}1.17$\times$ 
     & \cellcolor{gray!20}14.3 ms 
     & \cellcolor{gray!20}1.53$\times$ \\
    \hline
    \end{tabular}
    \caption{CodeGen-Mono 16B accuracy and latency on HumanEval with auto-regressive regular decoding (RD)  with DeepSpeed (DS) and vLLM, and \nameshort{}. Temperature is 0.2, nucleus top $p$ is 0.95, and draft model is CodeGen-Mono 350M.}
    \label{tab:codegen}
    \end{table*}

\subsection{Performance on code generation}

Table~\ref{tab:codegen} shows the accuracy and latency results of the CodeGen-Mono 16B model on the HumanEval task, with CodeGen-Mono 350M as the draft model.
The trends are similar to Table~\ref{tab:opt}: accuracy is neutral between regular decoding and speculative decoding; the first finished sequence is sped up by up to 2.65$\times$, representing a significant improvement in user-perceived latency; the average latency of all sequences is reduced by up to 2.43$\times$; the latency divergence between the first and last finished sequences increases with batch size.

Unlike Table~\ref{tab:opt}, the accuracy metric in Table~\ref{tab:codegen} increases with batch size: it is the percentage of examples where at least one correct generation exists in the batch. It represents an accuracy benefit of batched over single-sequence speculative decoding and more results will be presented in Section~\ref{sec:bene}.

Overall the speed-up ratios in Table~\ref{tab:codegen} are less than those in Table~\ref{tab:opt}, and we hypothesize that the main factor is the larger size of the draft model. 

\begin{table*}[h!]
    \centering
    \small
    \begin{tabular}{cclcrrrrrr}
    \hline
    \multirow{2}{*}{Prec.} & \multirow{2}{*}{Batch} & \multirow{2}{*}{Method} & \multirow{2}{*}{Pass@Batch} & \multicolumn{6}{c}{Mean per-token latency \& Speedup}\\
    \cline{5-10}
    &&&& \multicolumn{2}{c}{First} & \multicolumn{2}{c}{Last} & \multicolumn{2}{c}{All} \\
    \hline
    \multirow{10}{*}{BF16}
    & \multirow{2}{*}{1} & \reg{}& 36.6\% & 14.4 ms & 1$\times$ & 14.4 ms & 1$\times$ & 14.4  ms & 1$\times$ \\
    &  
        & \cellcolor{gray!20} \ours{}    
        & \cellcolor{gray!20} 34.1\% 
        & \cellcolor{gray!20} 4.6 ms 
        & \cellcolor{gray!20} 3.10$\times$ 
        & \cellcolor{gray!20} 4.6 ms 
        & \cellcolor{gray!20} 3.10$\times$ 
        & \cellcolor{gray!20} 4.6 ms 
        & \cellcolor{gray!20} 3.10$\times$ \\
    \cline{2-10}
    & \multirow{2}{*}{2} & \reg{}& 45.7\% & 14.6 ms & 1$\times$ & 14.6 ms & 1$\times$ & 14.6 ms & 1$\times$ \\
    &  
        & \cellcolor{gray!20} \ours{}    
        & \cellcolor{gray!20} 45.1\% 
        & \cellcolor{gray!20} 4.6 ms 
        & \cellcolor{gray!20} 3.16$\times$ 
        & \cellcolor{gray!20} 5.3 ms 
        & \cellcolor{gray!20} 2.74$\times$ 
        & \cellcolor{gray!20} 5.0 ms 
        & \cellcolor{gray!20} 2.94$\times$ \\
    \cline{2-10}
    & \multirow{2}{*}{4} & \reg{}& 48.8\% & 15.1 ms & 1$\times$ & 15.1 ms & 1$\times$ & 15.1 ms & 1$\times$ \\
    &  
        & \cellcolor{gray!20} \ours{}    
        & \cellcolor{gray!20} 51.8\% 
        & \cellcolor{gray!20} 4.7 ms 
        & \cellcolor{gray!20} 3.23$\times$ 
        & \cellcolor{gray!20} 7.0 ms 
        & \cellcolor{gray!20} 2.17$\times$ 
        & \cellcolor{gray!20} 5.7 ms 
        & \cellcolor{gray!20} 2.64$\times$ \\
    \cline{2-10}
    & \multirow{2}{*}{8} & \reg{}& 55.5\% & 16.0 ms & 1$\times$ & 16.0 ms & 1$\times$ & 16.0 ms & 1$\times$ \\
    &  
        & \cellcolor{gray!20} \ours{}    
        & \cellcolor{gray!20} 53.7\% 
        & \cellcolor{gray!20} 5.5 ms 
        & \cellcolor{gray!20} 2.92$\times$ 
        & \cellcolor{gray!20} 9.1 ms 
        & \cellcolor{gray!20} 1.75$\times$ 
        & \cellcolor{gray!20} 7.1 ms 
        & \cellcolor{gray!20} 2.25$\times$ \\
    \cline{2-10}
    & \multirow{2}{*}{16} & \reg{}& 59.1\% & 16.9 ms & 1$\times$ & 16.9 ms & 1$\times$ & 16.9 ms & 1$\times$ \\
    &  
        & \cellcolor{gray!20} \ours{}    
        & \cellcolor{gray!20} 57.9\% 
        & \cellcolor{gray!20} 7.3 ms 
        & \cellcolor{gray!20} 2.31$\times$ 
        & \cellcolor{gray!20} 13.0 ms 
        & \cellcolor{gray!20} 1.31$\times$ 
        & \cellcolor{gray!20} 9.6 ms 
        & \cellcolor{gray!20} 1.77$\times$ \\
    \hline
    \multirow{10}{*}{INT8}
    & \multirow{2}{*}{1} & \reg{}& 34.8\% & 11.0 ms & 1$\times$ & 11.0 ms & 1$\times$ & 11.0 ms & 1$\times$ \\
    &  
        & \cellcolor{gray!20} \ours{}    
        & \cellcolor{gray!20} 36.6\% 
        & \cellcolor{gray!20} 3.7 ms 
        & \cellcolor{gray!20} 2.99$\times$ 
        & \cellcolor{gray!20} 3.7 ms 
        & \cellcolor{gray!20} 2.99$\times$ 
        & \cellcolor{gray!20} 3.7 ms 
        & \cellcolor{gray!20} 2.99$\times$ \\
    \cline{2-10}
    & \multirow{2}{*}{2} & \reg{}& 40.9\% & 11.3 ms & 1$\times$ & 11.3 ms & 1$\times$ & 11.3 ms & 1$\times$ \\
    &  
        & \cellcolor{gray!20} \ours{}    
        & \cellcolor{gray!20} 43.3\% 
        & \cellcolor{gray!20} 3.7 ms 
        & \cellcolor{gray!20} 3.03$\times$ 
        & \cellcolor{gray!20} 4.4 ms 
        & \cellcolor{gray!20} 2.59$\times$ 
        & \cellcolor{gray!20} 4.1 ms 
        & \cellcolor{gray!20} 2.79$\times$ \\
    \cline{2-10}
    & \multirow{2}{*}{4} & \reg{}& 46.3\% & 11.8 ms & 1$\times$ & 11.8 ms & 1$\times$ & 11.8 ms & 1$\times$ \\
    &  
        & \cellcolor{gray!20} \ours{}    
        & \cellcolor{gray!20} 47.6\% 
        & \cellcolor{gray!20} 4.1 ms 
        & \cellcolor{gray!20} 2.84$\times$ 
        & \cellcolor{gray!20} 5.7 ms 
        & \cellcolor{gray!20} 2.07$\times$ 
        & \cellcolor{gray!20} 4.8 ms 
        & \cellcolor{gray!20} 2.44$\times$ \\
    \cline{2-10}
    & \multirow{2}{*}{8} & \reg{}& 51.2\% & 12.3 ms & 1$\times$ & 12.3 ms & 1$\times$ & 12.3 ms & 1$\times$ \\
    &  
        & \cellcolor{gray!20} \ours{}    
        & \cellcolor{gray!20} 55.5\% 
        & \cellcolor{gray!20} 4.5 ms 
        & \cellcolor{gray!20} 2.73$\times$ 
        & \cellcolor{gray!20} 7.5 ms 
        & \cellcolor{gray!20} 1.66$\times$ 
        & \cellcolor{gray!20} 5.8 ms 
        & \cellcolor{gray!20} 2.15$\times$ \\
    \cline{2-10}
    & \multirow{2}{*}{16} & \reg{}& 57.3\% & 13.6 ms & 1$\times$ & 13.6 ms & 1$\times$ & 13.6 ms & 1$\times$ \\
    &  
        & \cellcolor{gray!20} \ours{}    
        & \cellcolor{gray!20} 57.3\% 
        & \cellcolor{gray!20} 6.3 ms 
        & \cellcolor{gray!20} 2.16$\times$ 
        & \cellcolor{gray!20} 10.6 ms 
        & \cellcolor{gray!20} 1.29$\times$ 
        & \cellcolor{gray!20} 8.0 ms 
        & \cellcolor{gray!20} 1.70$\times$ \\
    \hline
    \end{tabular}
    \caption{A 7.8B code model's accuracy and latency on HumanEval with regular decoding (RD) with DeepSpeed (DS) and vLLM, and \nameshort{}. Temperature is 0.2, nucleus top $p$ is 0.95, and draft model is the first in Table~\ref{tab:drafts}.}
    \label{tab:inhouse}
    \end{table*}

Table~\ref{tab:inhouse} shows the accuracy and latency results of a custom 7.8B model, which was trained on text and code, on the HumanEval task, with a 310M-size draft model which is the first in Table~\ref{tab:drafts}.
The overall trends are similar to those in Table~\ref{tab:codegen} except that the speed-up ratios are substantially higher: the first finished sequence is sped up by up to 3.23$\times$, and the average latency of all sequences is reduced by up to 2.94$\times$.
We hypothesize that the draft model architecture choice is the main reason and we will look at impact of draft model designs next.

\subsection{Impact of draft model choices}

\label{sec:draft}
Table~\ref{tab:drafts} compares three draft models, all GPT2-like models with different architecture parameters as listed in the first three rows.
They are trained with the same data and for the same amount of tokens.
According to the fifth row, i.e., their stand-alone accuracy performance on HumanEval, the second draft model is more performant, likely due to its greater depth.
This is also supported by the sixth row which shows the chance of a draft token getting accepted during speculative decoding, and indeed the second draft model aligns better with the main model.
However, because the second draft model itself takes higher latency to generate draft tokens, the overall latency of speculative decoding is increased despite accepting more draft tokens.

Table~\ref{tab:optdrafts} is a similar comparison between two OPT draft models.
Surprisingly OPT 350M is worse than OPT 125M in both stand-alone ROUGE-2 score on XSum and token acceptance rate which represents worse alignment with the main model of OPT 13B.

\begin{table}[t]
\centering
\small
\begin{tabular}{l|l|c|c|c}
\hline
\multicolumn{2}{l|}{draft model} & A  & B & C \\
\hline
\multicolumn{2}{l|}{\#layer} & 4  & 8 & 4 \\
\multicolumn{2}{l|}{\#head}  & 16 & 16 & 32 \\
\multicolumn{2}{l|}{hidden dimension} & 2048 & 2048 & 4096 \\
\multicolumn{2}{l|}{\#param} & 310M & 510M & 1B \\
\multicolumn{2}{l|}{HumanEval pass@1} & 5.1\% & 11.4\% & 5.8\% \\
\multicolumn{2}{l|}{token acceptance rate} & 87.4\% & 88.5\% & 87.2\% \\
\hline
        & batch size 1  & 1.9 & 2.6 & 2.5 \\
draft   & batch size 2  & 2.0 & 2.6 & 2.6 \\
PTL     & batch size 4  & 2.0 & 2.7 & 2.7 \\
(ms)    & batch size 8  & 2.0 & 2.7 & 2.8 \\
        & batch size 16 & 2.1 & 3.0 & 3.1 \\
\hline
        & batch size 1  & 3.7 & 4.5 & 4.5 \\
1st Seq & batch size 2  & 3.7 & 4.4 & 4.8 \\
PTL     & batch size 4  & 4.1 & 5.0 & 5.2 \\
(ms)    & batch size 8  & 4.5 & 5.9 & 6.0 \\
        & batch size 16 & 6.3 & 7.6 & 7.7 \\
\hline
\end{tabular}
\caption{Comparisons between three draft models. \emph{PTL} stands for per-token latency, and \emph{1st Seq PTL} stands for that of the first finished sequence with \nameshort{}.}
\label{tab:drafts}
\end{table}

\begin{table}[t]
\centering
\small
\begin{tabular}{l|l|c|c}
\hline
\multicolumn{2}{l|}{draft model} & A  & B \\
\hline
\multicolumn{2}{l|}{\#layer} & 12 & 24 \\
\multicolumn{2}{l|}{\#head}  & 12 & 16 \\
\multicolumn{2}{l|}{hidden dimension} & 768 & 1024 \\
\multicolumn{2}{l|}{\#param} & 125M & 350M \\
\multicolumn{2}{l|}{XSum ROUGE-2} & 0.023 & 0.015 \\
\multicolumn{2}{l|}{token acceptance rate} & 78.5\% & 76.3\% \\
\hline
        & batch size 1  & 3.1 & 6.9 \\
draft   & batch size 2  & 5.0 & 8.6 \\
PTL     & batch size 4  & 5.0 & 8.5 \\
(ms)    & batch size 8  & 5.1 & 8.9 \\
\hline
        & batch size 1  & 8.5 & 14.2 \\
1st Seq & batch size 2  & 7.8 & 14.7 \\
PTL     & batch size 4  & 8.2 & 15.7 \\
(ms)    & batch size 8  & 9.6 & 16.6 \\
\hline
\end{tabular}
\caption{Comparisons between two OPT draft models. \emph{PTL} stands for per-token latency, and \emph{1st Seq PTL} stands for that of the first finished sequence with \nameshort{}.}
\label{tab:optdrafts}
\end{table}

\subsection{Benefits of batched speculative decoding}
\label{sec:bene}

\begin{figure}
    \centering
        \includegraphics[width=\linewidth]{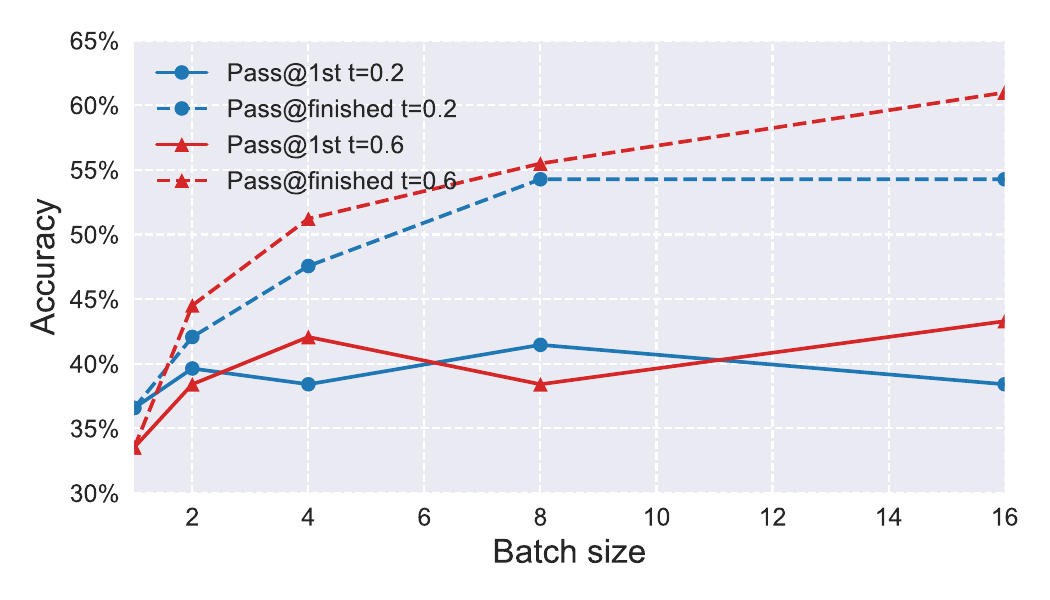}
        \caption{A 7.8B code model's accuracy on HumanEval with \nameshort{}, within a time budget of 2.5 seconds. $t$ is temperature.}
    \label{fig:passAtFirst}
\end{figure}

Figure~\ref{fig:passAtFirst} emulates a real-life application scenario where a service returns code recommendations to a user within a time budget. One of the recommendations is first displayed and the user has the option to flip through others. Ranking (here simply mean-logP based) is applied to pick the first displayed one. The Pass@First metric is the probability that the first displayed recommendation solves the problem correctly. The Pass@Finished metric is the probability that at least one of the finished recommendations within the time budget solves the problem correctly. These two metrics together quantify the accuracy quality of the service.

Figure~\ref{fig:passAtFirst} uses an end-to-end time budget of 2.5 seconds to generate 256-long sequences for any given prompt from HumanEval.
According to Table~\ref{tab:inhouse}, regular decoding under any setting would be unable to finish before time runs out, while single-sequence speculative decoding is able to return one recommendation and its Pass@First and Pass@Finished are the same and correspond to the left end point of the curves.
With BASS and as the batch size increases, Pass@Finished is increased up to 61\% and Pass@First is increased up to 43\% with a simple ranking strategy using model confidence of mean-logP value.
Both numbers are substantially higher than the mid-thirties accuracy by single-sequence speculative decoding.
A real-life application would use a domain-specific stopping criteria instead of a fixed length and a more sophisticated ranking method, but the relative comparisons among the competing methods are as captured by Figure~\ref{fig:passAtFirst} and \nameshort{} is clearly superior.

\subsection{Ablation studies}
\label{sec:ablation}

In Table~\ref{tab:ablation}, we compare latency  when certain alternative implementation choices are used.
With \nameshort{}-SPLIT, we launch per-sequence kernels to handle ragged tensors as illustrated in Figure~\ref{fig:method}(c).
The results suggest that the cost of launching more CUDA kernels in \nameshort{}-SPLIT out-weights the cost of wasted compute in \nameshort{}-PAD.
Note that this relation is task dependent and may change when the sequence lengths across a batch vary by a large margin.
For example, when applied on tasks of batch generation from a set of different prompts, the advantages of \nameshort{}-SPLIT could out-weights the cost.
With ``fixed draft size'', we use a constant draft length instead of Algorithm~\ref{alg:draftlength} that dynamically modifies draft length.
The results suggest that both the efficient attention calculation and the draft-length heuristic are important to the performance of \nameshort.

\begin{table}
\centering
\small
\begin{tabular}{lccc}
\hline
OPT 13B, XSum & \multicolumn{3}{c}{1st Seq PTL (ms)}\\
\cline{2-4}
\multicolumn{1}{r}{batch} & 2 & 4 & 8\\
\hline
\nameshort{} & \textbf{7.8} & \textbf{8.2} & \textbf{9.6} \\
\nameshort{}-SPLIT & 8.6 & 9.2 & 11.3\\
fixed draft size 4 & 8.9 & 9.6 & 11.3 \\
fixed draft size 6 & 8.9 & 9.0 & 10.2 \\
fixed draft size 8 & 8.9 & 9.1 & 10.3 \\
\hline
CG 16B, HumanEval & \multicolumn{3}{c}{1st Seq PTL (ms)}\\
\cline{2-4}
\multicolumn{1}{r}{batch} & 2 & 4 & 8\\
\hline
\nameshort{} & 9.3 & 9.8 & \textbf{11.1} \\
\nameshort{}-SPLIT & 10.1 & 11.7 & 12.7 \\
fixed draft size 4 & 9.7 & 10.2 & 12.3 \\
fixed draft size 6 & \textbf{9.1} & 9.7 & 11.6 \\
fixed draft size 8 & 9.7 & \textbf{9.5} & 13.1 \\
\hline
Code 7.8B, HumanEval & \multicolumn{3}{c}{1st Seq PTL (ms)}\\
\cline{2-4}
\multicolumn{1}{r}{batch} & 2 & 4 & 8\\
\hline
\nameshort{} & \textbf{3.7} & \textbf{4.1} & \textbf{4.5} \\
\nameshort{}-SPLIT & 4.0 & 4.4 & 5.2 \\
fixed draft size 4 & 4.6 & 5.1 & 6.3 \\
fixed draft size 6 & 4.3 & 4.9 & 5.8 \\
fixed draft size 8 & 4.0 & 4.3 & 5.1 \\
\hline
\end{tabular}
\caption{Ablation studies on latency impact of implementation choices. \emph{1st Seq PTL} is per-token latency of the first finished sequence. \nameshort{} is the default setting used in all other tables.}
\label{tab:ablation}
\end{table}

%% file: related.tex
\section{Related Work}

Efficient inference of LLMs has been a popular research topic in recent years. 
Model quantization techniques \cite{yao2022zeroquant, lin2023awq, frantar-gptq, fp8} employ lower-precision representations for model parameters (e.g., INT8, INT4, FP8) without significantly compromising accuracy. Pruning \cite{oneshotpruning} reduces memory footprints via sparsity.
Sparse attention techniques \citep{longformer,sparsetransformer} limit the number of tokens to attend to in order to reduce the complexity of attention layers, and thereby extend the maximum allowable sequence length.

Since its introduction, speculative decoding has seen numerous variations and improvements. Some proposals take a draft-model-free approach, by using an n-gram model to predict draft tokens \cite{fu2023lookahead}, or by using the model embedding to predict drafts \cite{cai2024medusa,li2024eagle}. SpecInfer \cite{miao2023specinfer} uses a {\it draft tree} to generate and organize multiple drafts for the main-model sequence in order to maximize the number of tokens accepted per step. \citet{su2023synergy} study the relation between batch size and the optimal fixed draft length for max throughput; it is however based on a primitive prototype implementation: rejected draft tokens are masked rather than discarded, which achieve sequence lengths that are uniform across a batch yet are unnecessarily large and inefficient.
The above works on speculative decoding are orthogonal to the discussions and ideas in this paper and can be combined. The conclusion presented in \cite{su2023synergy} may change with the kernel implementations in this paper.

%% file: ethic.tex
\section{Limitations}

This work, while advancing the state of the art, does not solve the efficient inference challenge of LLMs.
For example, GPU utilization during the context encoding phase of LLM inference can be over 70\% in our system, while the best achievable utilization during the incremental decoding phase is 15.8\% in this paper.
Although this is already significantly better than previous works, there is clearly substantial room to innovate and improve.

\section{Ethical Impact}

This work aims to increase the efficiency of deploying LLMs by utilizing the compute resources efficiently.
It can reduce carbon emissions associated with LLM deployment.
Additionally, driving down infrastructure costs can potentially encourage broader LLM adoption.
Impact of increased LLM usage on the society and associated risks are difficult to forecast.

%% file: appendix.tex
\appendix
\section{Appendix}
\label{sec:appendix}

\subsection{Quantization schemes and mechanisms}

This section describes the quantization schemes and kernels used for INT8 inference. Since granular assignment of precision improves the accuracy of quantized models, we assign the quantization ranges to the smallest granularity that allows us to compute the matrix multiplications in integer arithmetic, i.e., the granularity is set to the inner-product dimension. This translates to per-channel quantization for weights, dynamic per-token quantization for activation, and dynamic per-head and per-token quantization for keys, queries, and values.

\begin{figure}
    \centering
    \includegraphics[width=\columnwidth]{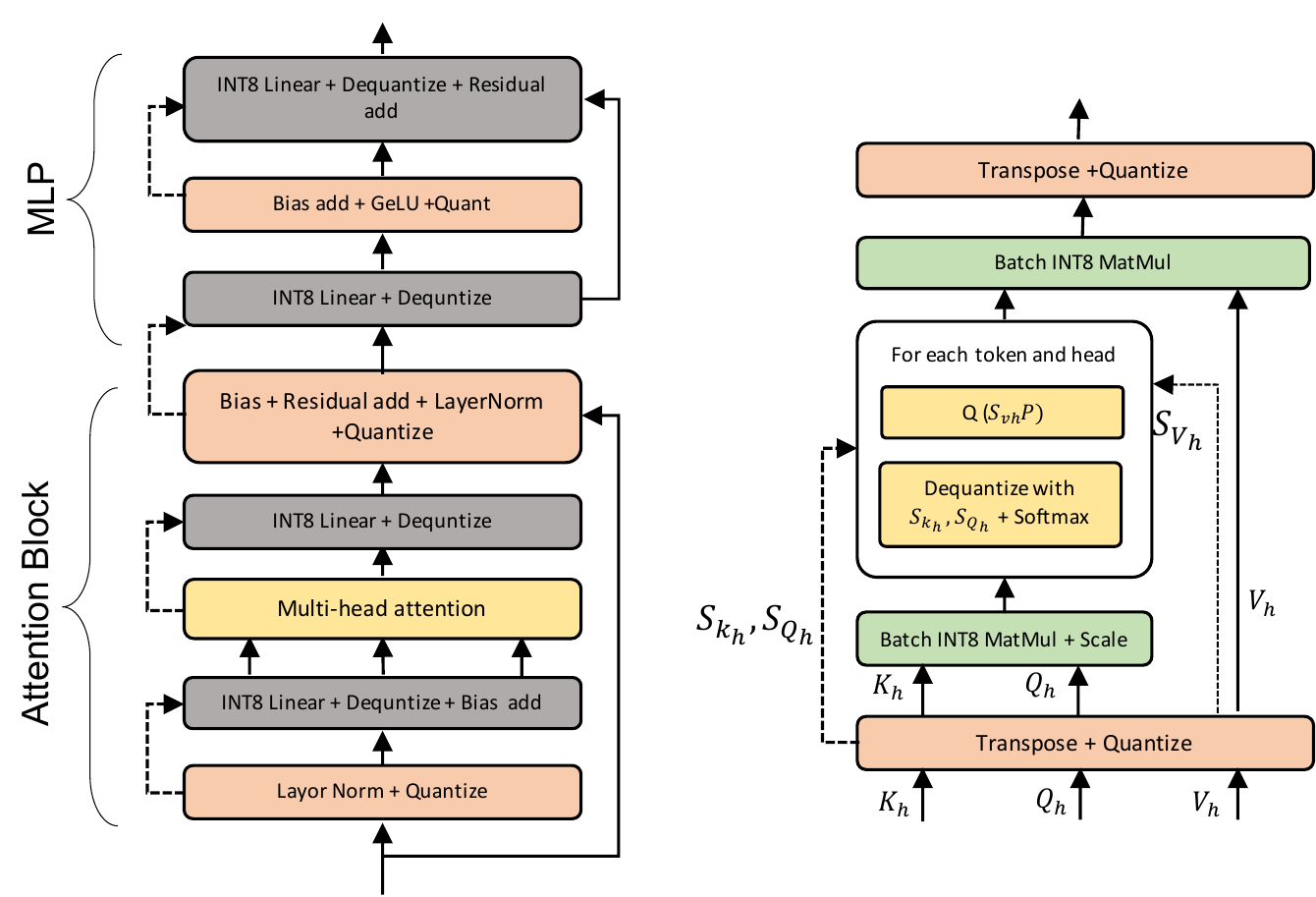}
    \caption{Inference data-flow with quantization}
    \label{fig:inference-flow}
\end{figure}

To mitigate the overheads associated with quantize and dequantize operations in the inference pipeline on a GPU, we have employed kernel fusion techniques as shown in Figure \ref{fig:inference-flow}. This involves amalgamating multiple operations to minimize CUDA kernel calls and memory fetches, thereby minimizing computation time.

Dynamic quantization can incur substantial overheads unless it is integrated with other operations in the inference pipeline. We fuse the quantize operation with layer-norm, GeLU, and transpose operations across the network. This approach eliminates the need for redundant memory reads, for example, reading the same data for layer-norm and quantization separately.

We use CUTLASS
INT8 GEMM kernels to fuse the dequantize operation and other element-wise computations, such as bias and residual additions with the GEMM operations. To optimize performance for small batch sizes, we adopt a strategy of pre-multiplying the weights and activation scales during the activation quantization operation, and subsequently retrieving them during the epilogue phase of the GEMM operations. The resulting fused GEMM blocks yield floating-point precision outputs, namely FP16, BF16 or FP32, depending on the selected format.

\subsection{Draft model choice and training}

For a draft model to be effective we need to maximize the number of tokens accepted by the main model while maintaining low per-token latency. Given the requirement, we have investigated architecture selection and observed that, at a fixed parameter count, having more attention heads (wider model) is better than having more layers with fewer attention heads (deeper model) since we have similar representation capabilities with wider-but-shallow models as narrow-but-deep models but at a much lower latency. We summarize the draft model architecture choices in Table \ref{tab:drafts}.

We trained each of these draft models using the same data for the main model with a context size of 2048 and a global 512 batch size across 8 servers each with 8 40GB Nvidia A100 GPUs. This translates to approximately 1 million tokens per batch. Using a learning rate of $3.5 \times 10^{-4}$ we train for 300k steps.
We use AdamW optimizer \cite{adamw} with $\beta_1 = 0.9$, $\beta_2 = 0.95$, and $\epsilon = 10^{-8}$. The warm-up steps were set to 2000, and a cosine annealing learning rate schedule was employed after reaching the peak learning rate.
The minimum learning rate was set to 10\% of the peak learning rate.
We use BF16 \cite{bf16} precision and set gradient clipping to 1.0 to enhance training stability.